\pdfoutput=1
\documentclass[11pt]{article}

\usepackage{acl}

\usepackage{times}
\usepackage{latexsym}

\usepackage[T1]{fontenc}

\usepackage[utf8]{inputenc}

\usepackage{microtype}

\usepackage{inconsolata}

\usepackage{graphicx}
\usepackage{algorithm}
\usepackage{amsfonts}
\usepackage{algpseudocode}
\usepackage{mathtools}
\usepackage{multirow}
\usepackage{booktabs}
\usepackage{tabularx}
\usepackage{array}
\usepackage{subcaption}
\usepackage[most]{tcolorbox}
\usepackage{listings}
\usepackage{xcolor}

\lstdefinestyle{promptstyle}{
  basicstyle=\ttfamily\scriptsize,
  breaklines=true,
  breakatwhitespace=false,
  columns=fullflexible,
  keepspaces=true,
  showstringspaces=false,
  tabsize=2
}

\newtcblisting{promptbox}{
  listing only,
  breakable,
  enhanced,
  colback=gray!3,
  colframe=gray!45,
  boxrule=0.35pt,
  arc=1pt,
  left=4pt,
  right=4pt,
  top=3pt,
  bottom=3pt,
  listing options={style=promptstyle}
}

\title{Routing by Reasoning Need: Trajectory-Aware Decoding Control for Diffusion Vision-Language Models}

\author{
 Yixiang Liu\textsuperscript{1}\thanks{\ \ Equal contribution.},
 Zhongxing Xu\textsuperscript{2}\footnotemark[1],
 Zhonghua Wang\textsuperscript{2}\footnotemark[1],
 Xiaoying Tang\textsuperscript{1}\thanks{\ \ Corresponding author.},
\\
 \textsuperscript{1}Southern University of Science and Technology,
 \textsuperscript{2}Monash University
\\
 \small{
   \textbf{Correspondence:} \href{mailto:tangxy@sustech.edu.cn}{tangxy@sustech.edu.cn}
 }
}

\begin{document}
\maketitle
\begin{abstract}

Diffusion vision-language models generate answers through iterative refinement, exposing intermediate answer trajectories that can be inspected and controlled at inference time.
However, this controllability creates a reasoning-need mismatch, where a universal generation length is applied to questions with different reasoning demands.
Visually closed questions may be harmed by continued refinement after a stable answer has formed, whereas reasoning-sensitive questions may be harmed by premature commitment.
We formulate this problem as reasoning-budget mismatch and study it in LLaDA-V. Rather than choosing a universal generation length, our training-free controller routes each example to early commitment, baseline preservation, or reasoning-supportive decoding using trajectory signals from answer closure, commitment evidence, and representation revision pressure, without using ground-truth answers. Across answer-focused, mixed-reasoning, and CoT-sensitive benchmarks, routed control improves robustness over fixed long decoding, pure short decoding, and single-rule interventions. The gains are not explained by shorter outputs alone. Answer-closed examples often benefit from commitment, whereas CoT-sensitive examples require preserving or supporting intermediate reasoning. Taken together, these results suggest diffusion VLM decoding should route inference-time control by the state suggested by the observed trajectory instead of relying on a universal decoding length.
\end{abstract}

\section{Introduction}
\begin{figure}[t]
  \centering
  \includegraphics[width=\columnwidth]{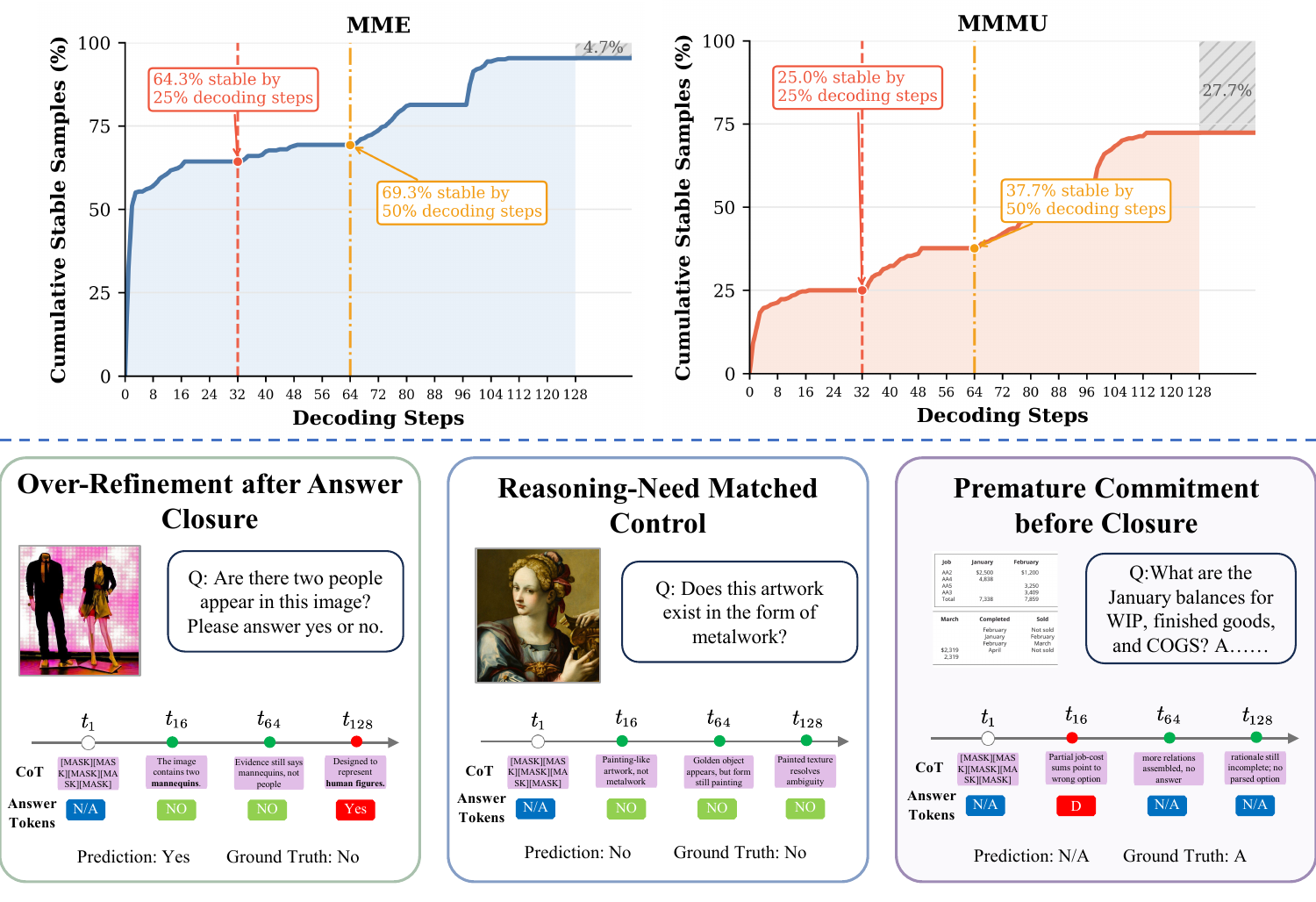}
  \caption{Motivation for reasoning-need matched decoding control. The top row shows answer-closure timing under the fixed 128-step decoding budget across benchmark regimes. Gray regions indicate decoding intervals where no valid answer has converged within the 128-step trajectory. The bottom row illustrates three representative trajectory outcomes, including over-refinement after closure, matched control, and premature commitment before closure. These patterns motivate routing the decoding action by the state suggested by the observed trajectory instead of applying a universal generation length.}
  \label{fig:motivation}
\end{figure}
Diffusion vision-language models (VLMs) change what can be controlled during decoding. In left-to-right autoregressive generation, the model commits to a token sequence one position at a time. After a token is emitted, later computation can condition on it but cannot directly revise it. Diffusion language models instead generate through iterative denoising or masked refinement, and recent diffusion VLMs extend this design to multimodal instruction following \citep{llada2026,lladav2025}. This makes decoding a sequence of observable revisions over a partially specified answer rather than a single irreversible trace. Because these intermediate answer trajectories are visible at inference time, they can indicate whether a candidate answer has stabilized, whether the trajectory provides sufficient grounding-and-format evidence for commitment, and whether the model is still changing the representation that will support the final answer. For a visually closed VQA example, such as many compact diagnostic questions in MME \citep{mme}, the answer may appear early and remain consistent across refinements \citep{lidiffusion, xu2026thinking}. 
By contrast, for geometry, chemistry, or multi-hop reasoning examples, such as MMMU-style expert questions \citep{mmmu}, the same early surface answer may be unreliable because the model has not yet organized the relevant relations. 
As shown in the top row of Figure~\ref{fig:motivation}, answer closure can occur early for compact visual questions, while reasoning-intensive examples may require later refinement before a reliable answer emerges. 
Diffusion decoding therefore exposes both the final answer and the revision process that leads to it.

This controllability also creates risk under a fixed decoding budget \citep{li2025beyond}. In our experiments, the default fixed-budget reference uses a 128-step refinement trajectory, which assigns every sample the same amount of decoding computation even when questions differ sharply in their reasoning demands. Visually closed questions may be harmed by continued refinement after a stable answer has formed, while reasoning-sensitive questions may be harmed by premature commitment. We call this failure a reasoning-budget mismatch. The mismatch is not simply that some outputs should be shorter and others longer. It is that different samples require different inference-time control actions because the uncertainty left after early refinement has different meanings. In answer-closed cases, the suitable action is early commitment. In unresolved cases, where intervention is not justified, the safer action is to preserve the fixed-budget baseline. In CoT-sensitive cases, where the final answer should remain supported by intermediate reasoning, the suitable action is reasoning-supportive decoding rather than premature commitment.

Figure~\ref{fig:motivation} bottom illustrates this mismatch. Some trajectories reach a stable answer early and are later degraded by additional refinement. Some trajectories remain stable under continued decoding and should be preserved. Others do not yet contain a reliable answer or support-bearing rationale, so early commitment would truncate the reasoning process. These cases suggest that the central question is not whether diffusion decoding should be globally short or long, but which control action is appropriate for the current trajectory.

This paper studies reasoning-budget mismatch in LLaDA-V. Rather than choosing a universal generation length, we propose a training-free inference-time controller that routes each example to one of three actions: early commitment, baseline preservation, or reasoning-supportive decoding. The controller uses trajectory signals computed without ground truth from answer closure, commitment evidence, and representation revision pressure. These signals do not estimate answer correctness or define a ground-truth semantic label of the example. Instead, we use reasoning need as an operational decoding concept: it denotes the control action suggested by the observed diffusion trajectory. This distinction explains why fixed long decoding, pure short decoding, and single-rule interventions can each help in some regimes but fail in others. Diffusion VLM decoding should route inference-time control by the state suggested by the observed trajectory rather than relying on a universal decoding length.

Our contributions are threefold.

\begin{itemize}
\item We identify reasoning-budget mismatch in diffusion VLM decoding, where the same refinement budget can over-refine answer-closed examples while under-supporting reasoning-sensitive examples.

\item We propose a training-free routed decoding controller for LLaDA-V, which selects among early commitment, baseline preservation, and reasoning-supportive decoding using trajectory signals computed without ground-truth answers.

\item We evaluate the controller across answer-focused, mixed-reasoning, and CoT-sensitive settings, showing that routing inference-time control actions is more robust than applying a single global length rule.
\end{itemize}

\section{Related Work}

\noindent\textbf{Diffusion VLMs and adaptive decoding.}
Diffusion language models generate text through iterative denoising or masked refinement rather than left-to-right token commitment \citep{llada2026}. LLaDA-V extends this formulation to multimodal instruction following and visual reasoning, making intermediate states observable during vision-language generation \citep{lladav2025}. This process provides a natural interface for inference-time control, since the decoder can inspect partial predictions, uncertainty patterns, and refinement dynamics before the final answer is fixed. Recent adaptive decoding methods exploit such internal signals by adjusting denoising steps, block granularity, token representations, or reasoning modes \citep{fastdllm,adablock,dllmvar,xu2026thinking,li2025beyond}. These methods show that fixed decoding schedules are often suboptimal, but their objectives are usually efficiency, uncertainty reduction, or hallucination mitigation. Our work studies a different mismatch. The appropriate control action depends on the sample's trajectory state. Some samples are already answer-closed and benefit from early commitment, some remain unresolved, and others require reasoning-supportive decoding rather than shorter decoding.

\noindent\textbf{Reasoning length, grounding, and CoT sensitivity.}
Longer reasoning is not uniformly beneficial. In multimodal settings, extended reasoning can improve task solving while weakening visual grounding, and reasoning-induced hallucination should be distinguished from perception-induced errors \citep{xu2026more,mirage2025}. CoT prompting is task-conditional. It helps when intermediate reasoning supports the decision, but can be unnecessary or harmful when the answer is already recoverable from the input \citep{tocotornot,cheng2025cotobscures,cotfaithfulness2025}. This suggests that reasoning should be instance-conditional rather than globally enabled or globally suppressed. The relevant decision is not whether to use more or less reasoning for all samples, but whether the current trajectory still requires structured reasoning support. Grounded reasoning methods show that when reasoning is necessary, the better remedy is often to anchor intermediate steps in visual evidence rather than truncate the chain \citep{gcot2025}. Our distinction between answer-closed and CoT-sensitive cases follows this view. Early commitment is useful when further refinement mainly adds drift, while preserved or structured reasoning is useful when the decision still depends on unresolved visual-textual evidence.

\noindent\textbf{Decoding-time control and its boundary.}
Training-free decoding-time methods modify generation without retraining. Representative approaches include visual contrastive decoding, over-trust penalties, layer-contrastive decoding, self-corrective decoding, and memory-based visual refinement \citep{leng2024vcd,huang2024opera,chuang2024dola,sid,memvr}. Entropy-aware decoding further shows that token-level uncertainty can guide mode switching and visual-anchor injection during uncertain reasoning states \citep{xu2026more}. These methods demonstrate that internal generation signals can support effective test-time intervention. Our results suggest a boundary for single-rule interventions. Pure short decoding, fixed long decoding, and revision-based control each help in some regimes and fail in others. We therefore treat internal signals not as universal hallucination or factuality detectors, but as evidence for choosing among control actions. This reframes diffusion VLM decoding as a regime-conditioned decision problem, where the system must decide whether to commit early, preserve the baseline path, or apply reasoning-supportive decoding.

\begin{figure*}[t]
    \centering
    \includegraphics[width=\textwidth]{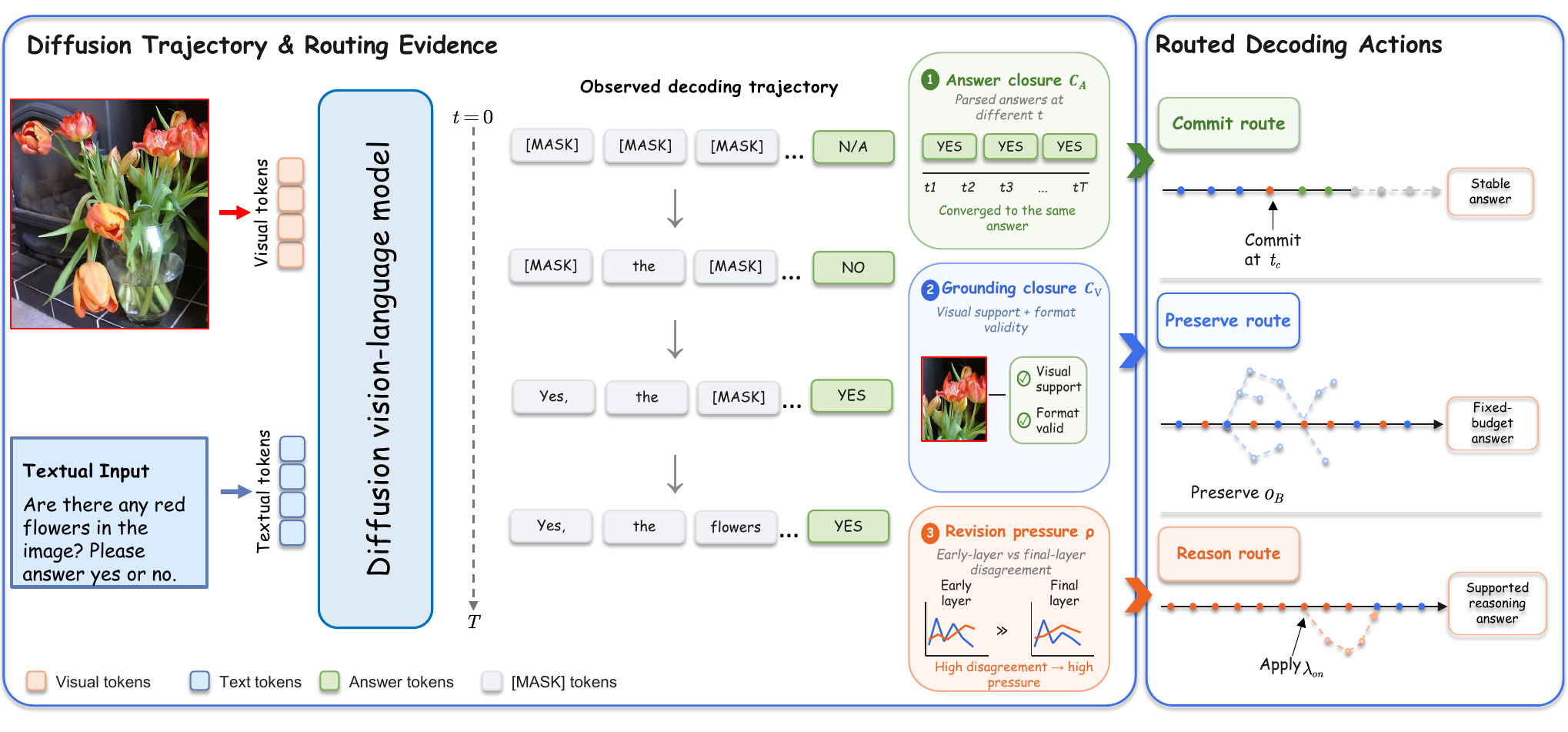}
    \caption{
    Overview of reasoning-need routed decoding control.
    }
    \label{figframework}
\end{figure*}
\section{Method}
Figure~\ref{figframework} gives an overview of our reasoning-need routed decoding framework. 
The controller observes a fixed-budget diffusion trajectory and extracts three signals from intermediate refinement states, including answer closure, grounding-and-format evidence from format and visual-engagement guards, and representation revision pressure. These signals do not use gold labels or estimate correctness directly.
They operationalize the state suggested by the observed trajectory as the safest inference-time action under the observed trajectory, routing each example to early commitment, baseline preservation, or reasoning-supportive decoding instead of using a universal short or long budget.
\subsection{Problem Setup}

We study multimodal question answering with a frozen diffusion VLM. 
Each input is $x=(x_v,x_q,c,I)$, where $x_v$ is the image, $x_q$ is the question, $c$ denotes optional answer choices, and $I$ denotes an optional reasoning instruction. 
The model returns an answer $\hat a$ and, when requested, a rationale $\hat r$.

Let $B=(T,L,M)$ denote the decoding budget, where $T$ is the number of denoising steps, $L$ is the generation length, and $M$ is the block length. 
The fixed-budget reference uses the same budget for every sample. 
At denoising step $t$, the diffusion state is $z_t=(y_t,H_t,P_t)$, where $y_t$ is the token state, $H_t$ denotes hidden states, and $P_t$ denotes token distributions. 
We write the fixed-budget trajectory and parsed output as
\begin{equation}
\begin{aligned}
z_{t+1} &= F_\theta(z_t,x,B),\\
\tau_B(x) &= (z_0,\ldots,z_T),\\
o_B(x) &= \psi(y_T)=(\hat a_B,\hat r_B).
\end{aligned}
\label{eq:fixed-budget}
\end{equation}
Here $\theta$ is frozen and $\psi$ is the benchmark parser. 
Our controller does not update model weights, train a verifier, or use gold labels. 
It only selects an action from the decoding trajectory.

\subsection{Representative Trajectory Patterns}
\begin{figure*}[t]
    \centering
    \includegraphics[width=0.95\textwidth]{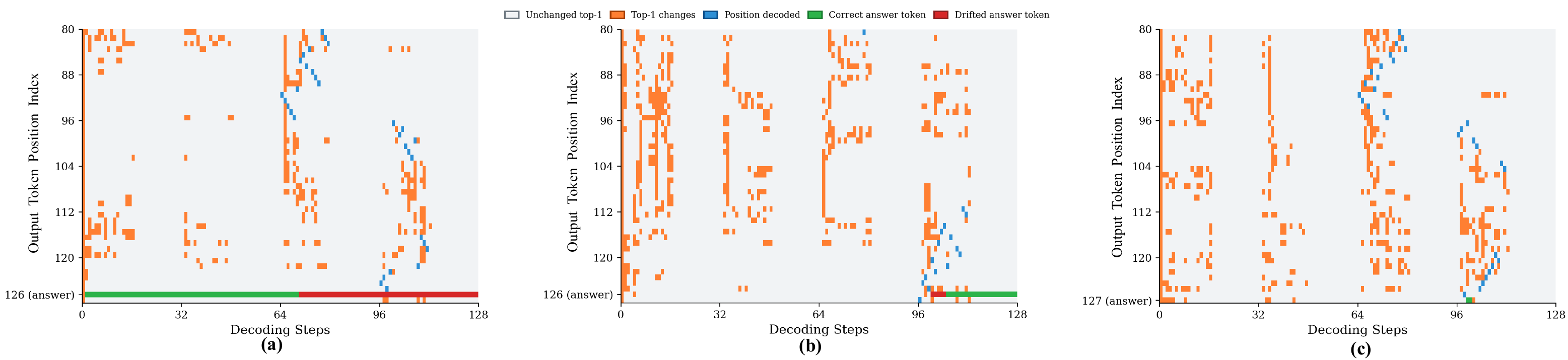}
    \caption{
    Representative answer-token trajectory patterns that motivate reasoning-need routing.
    Each panel shows top-1 token changes, decoded positions, and answer-token correctness across diffusion decoding steps.
    (a) An early-closure case where the correct answer appears early but later refinement drifts to an incorrect answer, motivating early commitment.
    (b) A late-converging case where the answer becomes reliable only after sufficient refinement, motivating baseline preservation when early commitment is not justified.
    (c) A reasoning-heavy unstable case where the answer trajectory remains unsettled under simple length control, motivating reasoning-supportive decoding rather than a universal short or long budget.
    }
    \label{fig:trajectory-patterns}
\end{figure*}

Figure~\ref{fig:trajectory-patterns} illustrates three answer-token trajectories that motivate our routing policy.
The plots show when output positions are decoded, where top-1 predictions change, and whether the answer token remains correct or drifts.
The examples correspond to three decoding states: early answer closure with later drift, late convergence that should be preserved, and reasoning-heavy instability that requires additional support.
These patterns motivate selecting among early commitment, baseline preservation, and reasoning-supportive decoding according to the sample's reasoning need, rather than using one fixed decoding length.

\subsection{Routing Evidence}

The router uses three trajectory signals: answer closure, commitment evidence, and representation revision pressure.

\paragraph{Answer closure.}
Answer closure measures whether the parsed answer has stabilized. 
Let $\psi_A(y_t)$ return the parsed answer from state $y_t$, or $\bot$ if no valid answer is available. 
For a routing window $\mathcal W$, let $\hat a_{\mathcal W}$ be the most frequent valid parsed answer in that window, with $\hat a_{\mathcal W}=\bot$ if no valid answer appears. 
We define
\begin{equation}
C_A(x)
=
\mathbb{1}[\hat a_{\mathcal W}\neq \bot]
\frac{1}{|\mathcal W|}
\sum_{t\in\mathcal W}
\mathbb{1}[\psi_A(y_t)=\hat a_{\mathcal W}].
\label{eq:answer-closure}
\end{equation}
High $C_A$ means that the same parseable answer repeatedly appears; it does not imply correctness.

\paragraph{visual-format closure.}
We use visual engagement as a proxy for whether the trajectory attends to image tokens, not whether the answer is true. 
The proxy aggregates parser validity, visual-attention evidence, prefix-dominance risk, and parsed-answer flip risk:
\begin{equation}
\begin{aligned}
\mathbf{s}_V(x)
&=
\Bigl(
C_{\mathrm{fmt}}(x), C_{\mathrm{vis}}(x),\\
&\quad
R_{\mathrm{prefix}}(x), R_{\mathrm{flip}}(x)
\Bigr),\\
C_V(x)
&=
g_V\!\left(\mathbf{s}_V(x)\right)\in[0,1].
\end{aligned}
\label{eq:grounding-format-closure}
\end{equation}
Here $C_{\mathrm{fmt}}$ checks answer-format validity, $C_{\mathrm{vis}}$ is computed from visual-attention mass over image-token positions, and $R_{\mathrm{prefix}}$ and $R_{\mathrm{flip}}$ capture prefix-dominance and answer-instability risks.
The fixed aggregation function $g_V$ provides only routing evidence and does not verify whether the answer is visually true.

\paragraph{Representation revision pressure.}
Representation revision pressure measures whether early and final layer distributions still disagree. 
Let $p_{e,t,j}$ and $p_{f,t,j}$ be token distributions from an early layer and the final layer at position $j$. 
For route positions $\Omega_t$, we compute
\begin{equation}
\begin{gathered}
J_t
=
\frac{1}{|\Omega_t|}
\sum_{j\in\Omega_t}
\mathrm{JS}\!\left(p_{e,t,j}\,\|\,p_{f,t,j}\right),\\
\rho(x)
=
\operatorname{Agg}_{t\in\mathcal W} J_t .
\end{gathered}
\label{eq:rrp}
\end{equation}
High $\rho(x)$ indicates that the model is still revising its internal distribution. 
It is a routing signal, not a hallucination detector or correctness score.

\subsection{Routing Policy}

The controller chooses among three actions, including \textsc{CommitRoute}, \textsc{PreserveRoute}, and \textsc{ReasonRoute}. 
Let $K(x)\in\{0,1\}$ be the fixed CoT-sensitive guard, set from the prompt, task format, or disclosed route configuration. 
Let $G_{\mathrm{on}}(x)\in\{0,1\}$ be the fixed route-on guard for the reported operating point. 
With thresholds fixed before evaluation, define
\begin{equation}
\begin{aligned}
U_R(x)
&\equiv
\rho(x)\ge \beta_\rho
\lor C_A(x)<\beta_A,\\
U_E(x)
&\equiv
C_A(x)\ge \alpha_A
\land C_V(x)\ge \alpha_V\\
&\quad\land \rho(x)\le \alpha_\rho .
\end{aligned}
\label{eq:route-predicates}
\end{equation}
We use $\beta_A \le \alpha_A$ and $\alpha_\rho \le \beta_\rho$, so intermediate cases fall back to preservation.
The reason and commit conditions are as follows.
\begin{equation}
\begin{aligned}
\mathcal R(x)
&=
K(x)\,G_{\mathrm{on}}(x)\,
\mathbf{1}[U_R(x)],\\
\mathcal E(x)
&=
\mathbf{1}[U_E(x)]\,
\chi_{\mathrm{commit}}(x).
\end{aligned}
\label{eq:route-conditions}
\end{equation}
Here $\chi_{\mathrm{commit}}(x)$ denotes fixed hard guards, such as parser availability, operating-point constraints, or route-specific exclusion rules. 
Signals already summarized by $C_V(x)$ are not counted again in $\chi_{\mathrm{commit}}$.

The routing policy is
\begin{equation}
\Pi(x)=
\begin{cases}
\mathsf{reason}, & \mathcal R(x)=1,\\
\mathsf{commit}, & \mathcal R(x)=0 \land \mathcal E(x)=1,\\
\mathsf{preserve}, & \text{otherwise}.
\end{cases}
\label{eq:routing-policy}
\end{equation}
The reason condition is evaluated before commitment because answer stability alone is insufficient in CoT-sensitive interfaces. 
When $\mathcal R(x)=1$, the controller avoids early commitment and applies the disclosed reasoning-supportive configuration. 
When neither reason nor commit is justified, the controller preserves the fixed-budget output.

\subsection{Routing Actions}

For \textsc{ReasonRoute}, the controller uses a fixed reasoning-supportive decoding configuration. 
Let $D_{\theta}(x;B,\lambda)$ denote the parsed output obtained by running the frozen model under budget $B$ and inference-time configuration $\lambda$. 
Then
\begin{equation}
o_r(x)=D_{\theta}(x;B,\lambda_{\mathrm{on}}),
\label{eq:reason-route}
\end{equation}
where $\lambda_{\mathrm{on}}$ is fixed and disclosed for the reported operating point.

For \textsc{CommitRoute}, the controller commits to the first locally stable parsed answer when the enabled closure guards fire. 
If an online commit step is used, let
\begin{equation}
t_c=\min\{t\mid \mathcal E_t(x)=1\},
\label{eq:first-commit}
\end{equation}
where $\mathcal E_t(x)$ applies the commit predicate in Eq.~\ref{eq:route-conditions} to the local routing window ending at step $t$. 
Let $\hat a_c=\psi_A(y_{t_c})$. 
The committed output is
\begin{equation}
o_c(x)=A_c(\hat a_c,\tau_B,o_B),
\label{eq:commit-route}
\end{equation}
where $A_c$ is the fixed commit implementation for the reported row. 
In the main online setting, $A_c$ fixes the answer field to the first stable parsed answer. 
Trace composition or source selection variants are marked as offline analyses.

\textsc{PreserveRoute} returns the fixed-budget output unchanged:
\begin{equation}
o_p(x)=o_B(x).
\label{eq:preserve-route}
\end{equation}

The final output is
\begin{equation}
o(x)=
\begin{cases}
o_c(x), & \Pi(x)=\mathsf{commit},\\
o_p(x), & \Pi(x)=\mathsf{preserve},\\
o_r(x), & \Pi(x)=\mathsf{reason}.
\end{cases}
\label{eq:routed-output}
\end{equation}

\subsection{Routed Decoding Algorithm}

Algorithm~\ref{alg:routed-decoding} summarizes the routing procedure.
The algorithm computes trajectory evidence, applies the route conditions in Eq.~\ref{eq:route-conditions}, and returns the corresponding routed output.
The routine $\mathrm{ObtainRoutingTrace}$ denotes the trace used for routing; rows that require cached candidate traces are marked separately as offline analyses and are not used for latency claims.

\begin{algorithm}[t]
\small
\caption{Reasoning-Need Routed Decoding Control}
\label{alg:routed-decoding}
\begin{algorithmic}[1]
\Require input $x=(x_v,x_q,c,I)$, budget $B$, frozen model $D_\theta$
\Ensure routed output $o$
\State $\tau_B,o_B \leftarrow \mathrm{ObtainRoutingTrace}_{\theta}(x,B)$
\State $C_A \leftarrow \mathrm{AnswerClosure}(\tau_B)$
\State $C_V \leftarrow \mathrm{GroundingFormatClosure}(\tau_B,x)$
\State $\rho \leftarrow \mathrm{RevisionPressure}(\tau_B)$
\State $K \leftarrow \mathrm{CoTSensitiveGuard}(x)$
\State $G_{\mathrm{on}} \leftarrow \mathrm{RouteOnGuard}(x,\tau_B)$
\State $\chi_{\mathrm{commit}} \leftarrow \mathrm{CommitGuard}(x,\tau_B)$
\State Compute $U_R,U_E,\mathcal R,\mathcal E$ using Eqs.~\ref{eq:route-predicates}--\ref{eq:route-conditions}
\If{$\mathcal R(x)=1$}
    \State \Return $D_{\theta}(x;B,\lambda_{\mathrm{on}})$
\ElsIf{$\mathcal E(x)=1$}
    \State $t_c \leftarrow \mathrm{FirstCommitStep}(\tau_B)$
    \State $\hat a_c \leftarrow \psi_A(y_{t_c})$
    \State \Return $A_c(\hat a_c,\tau_B,o_B)$
\Else
    \State \Return $o_B$
\EndIf
\end{algorithmic}
\end{algorithm}





\section{Experiments}

\subsection{Experimental Setup}

We evaluate LLaDA-V under answer-focused, mixed-reasoning, and CoT or support-sensitive settings. 
MME \citep{mme}, MMMU \citep{mmmu}, and MMStar \citep{mmstar} use compact final-answer interfaces, but differ in reasoning demands. 
MME contains many compact visual diagnostic questions, while MMMU and MMStar include more expert-level, fine-grained, or vision-indispensable cases. 
ScienceQA-IMG \citep{scienceqa}, A-OKVQA \citep{aokvqa}, and MME-CoT \citep{mmecot} are used as support-sensitive settings. 
ScienceQA-IMG provides annotated explanations, A-OKVQA requires image-grounded commonsense and world knowledge, and MME-CoT directly evaluates multimodal chain-of-thought quality.

All methods use the same frozen LLaDA-V model and the original benchmark parser. 
We compare fixed-budget baselines with different generation budgets, including Len2, Len32, Len64, and Len128, where Len128 is the default fixed-budget reference. 
We also compare visual and contrastive decoding controls, adaptive-budget controls, and our routed controller when benchmark-matched artifacts are available. 
Final-answer accuracy is the primary metric for benchmark comparison. 
For auxiliary analysis, we report decoding budget statistics in Appendix and CoT rationale diagnostics on the ScienceQA-IMG active-control subset. 
These diagnostics use reference-based text metrics and GPT-5.5 judge scores, but are not treated as gold-standard human faithfulness evidence.



\newcommand{\MainRegimeResultsTable}{%
\begin{table*}[t]
\centering
\small
\setlength{\tabcolsep}{3.0pt}
\renewcommand{\arraystretch}{1.10}
\begin{tabularx}{\textwidth}{@{}l*{6}{>{\centering\arraybackslash}X}@{}}
\toprule
& \multicolumn{3}{c}{\textbf{Answer-Focused Evaluation}}
& \multicolumn{3}{c}{\textbf{CoT-Sensitive Evaluation}} \\
\cmidrule(lr){2-4}\cmidrule(l){5-7}
Method
& MME
& MMMU
& MMStar
& \shortstack{SQA-\\IMG}
& \shortstack{A-OKVQA}
& \shortstack{MME-\\CoT} \\
\midrule

\multicolumn{7}{@{}l}{\textit{Global budget and task baselines}} \\
\midrule
Len2
& 79.56
& 48.89
& 46.88
& 84.28
& 78.75
& 48.72 \\
Len32
& 73.81
& 28.78
& 25.00
& 53.00
& 82.50
& 49.57 \\
Len64
& 74.60
& 36.11
& 17.19
& 76.65
& 82.50
& 49.57 \\
Len128
& 73.41
& 42.44
& 49.90
& 75.31
& 83.75
& 49.57 \\

\midrule
\multicolumn{7}{@{}l}{\textit{Visual and contrastive controls}} \\
\midrule
VCD
& 73.41
& 47.33
& 32.81
& 80.80
& 83.75
& 49.29 \\
SID
& 68.65
& 47.78
& 34.38
& 81.00
& 82.50
& 50.43 \\
MEMVR
& --
& 48.56
& 35.94
& --
& --
& 19.37 \\

\midrule
\multicolumn{7}{@{}l}{\textit{Adaptive-budget controls}} \\
\midrule
Fast-dLLM
& 72.33
& 42.89
& 34.38
& 75.61
& 83.75
& 49.57 \\
AdaBlock
& 71.40
& 42.33
& 35.94
& 74.81
& 80.00
& 49.29 \\
DAEDAL-lite
& 72.03
& 42.44
& 32.81
& 75.90
& 83.75
& 49.57 \\
dLLM-Var
& 69.29
& 40.89
& 20.31
& 70.20
& 85.00
& 48.15 \\

\midrule
\multicolumn{7}{@{}l}{\textit{Reasoning-need routed control}} \\
\midrule
\textbf{Ours}
& \textbf{79.96}
& \textbf{49.44}
& \textbf{50.77}
& \textbf{88.60}
& \textbf{88.75}
& \textbf{52.42} \\

\bottomrule
\end{tabularx}
\caption{Controlled results across answer-focused and CoT-sensitive evaluation settings. Scores are accuracies. Missing entries indicate that no benchmark-matched artifact was available.}
\label{tab:controlled-regime-results}
\end{table*}
}

\newcommand{\EfficiencyBudgetTable}{%
\begin{table}[t]
\centering
\scriptsize
\setlength{\tabcolsep}{3.2pt}
\renewcommand{\arraystretch}{1.12}
\resizebox{\columnwidth}{!}{%
\begin{tabular}{llccc}
\toprule
Benchmark & Method & Steps & Time & Len. \\
\midrule
\multicolumn{5}{@{}l}{\textbf{Answer-closed / mixed settings}} \\
\midrule
MME & Len128 & 128.00 & 22.53s & 47.0 \\
MME & Len64 & 64.00 & 10.60s & 23.43 \\
MME & Len32 & 32.00 & 4.85s & 11.40 \\
MME & Len2 & \textbf{2.00} & \textbf{0.28s} & 1.0 \\
MME & Ours & 128.00 & 22.77s & 40.01 \\
\midrule
\multicolumn{5}{@{}l}{\textbf{CoT-sensitive setting}} \\
\midrule
MME-CoT & Len128 & 95.91 & 25.52s & 89.42 \\
MME-CoT & Len64 & 40.05 & 10.41s & 45.05 \\
MME-CoT & Len32 & 28.09 & 4.97s & 23.55 \\
MME-CoT & Len2 & \textbf{2.00} & 0.35s & \textbf{1.46} \\
MME-CoT & Ours & 102.09 & 25.67s & 89.26 \\
\midrule
ScienceQA-IMG & Len128 & 101.47 & 24.6s & 99.94 \\
ScienceQA-IMG & Len64 & 64.00 & 11.10s & 23.57 \\
ScienceQA-IMG & Len32 & 32.00 & 5.64s & 11.15 \\
ScienceQA-IMG & Len2 & \textbf{2.00} & 0.31s & \textbf{1.00} \\
ScienceQA-IMG & Ours & 118.09 & 24.89s & 110.33 \\
\bottomrule
\end{tabular}
}
\caption{Budget and efficiency controls. The table reports decoding budget rather than answer accuracy. Entries marked with  are approximate estimates from measured budget trends when full audit latency or output length was unavailable. Len2 is inexpensive for answer-closed settings, whereas the MME-CoT route retains a long reasoning budget, indicating that its gain is not an output-shortening effect.}
\label{tab:efficiency-budget}
\end{table}
}

\subsection{Controlled Results Across Regimes}

\MainRegimeResultsTable

Table~\ref{tab:controlled-regime-results} summarizes controlled comparisons across answer-focused and support-sensitive settings. 
The results support three observations.

First, fixed long decoding is not consistently optimal in answer-focused settings. 
On MME, routed control reaches 79.96, improving over Len128 by 6.55 points and slightly exceeding Len2 at 79.56. 
On MMMU, it reaches 49.44, improving over Len128 by 7.00 points and exceeding Len2 at 48.89. 
These gains suggest that routed control is not simply applying a global short budget. 
Instead, it commits when the trajectory is low-risk and preserves the baseline when early commitment is not justified. 
On MMStar, routed control reaches 50.77, slightly above Len128 at 49.90, which further supports sample-wise action selection in compact-answer multimodal settings.

Second, final-answer closure can appear even in rationale-bearing benchmarks, but this should not be interpreted as absence of reasoning. 
On ScienceQA-IMG, routed control reaches 88.60, improving over Len128 by 13.29 points and over Len2 by 4.32 points. 
Because ScienceQA-IMG provides explanations but Table~\ref{tab:controlled-regime-results} reports final-answer accuracy, this result shows that final-answer scoring can exhibit early answer closure within a CoT-capable benchmark. 
It does not imply that the underlying task lacks reasoning structure.

Third, pure short decoding is not CoT-safe. 
On MME-CoT, Len2 falls to 48.72, below Len128 at 49.57, while routed control reaches 52.42, suggesting that premature commitment can remove useful reasoning structure. 
On targeted A-OKVQA closure, routed control reaches 88.75, outperforming Len128 at 83.75 and the strongest non-routed baseline at 85.00. 
These patterns support trajectory-aware routing over a universal length policy: short decoding helps stable answers, long decoding preserves reasoning capacity, and routing selects the safer action.


\begin{figure}[t]
\centering
\includegraphics[width=0.49\columnwidth]{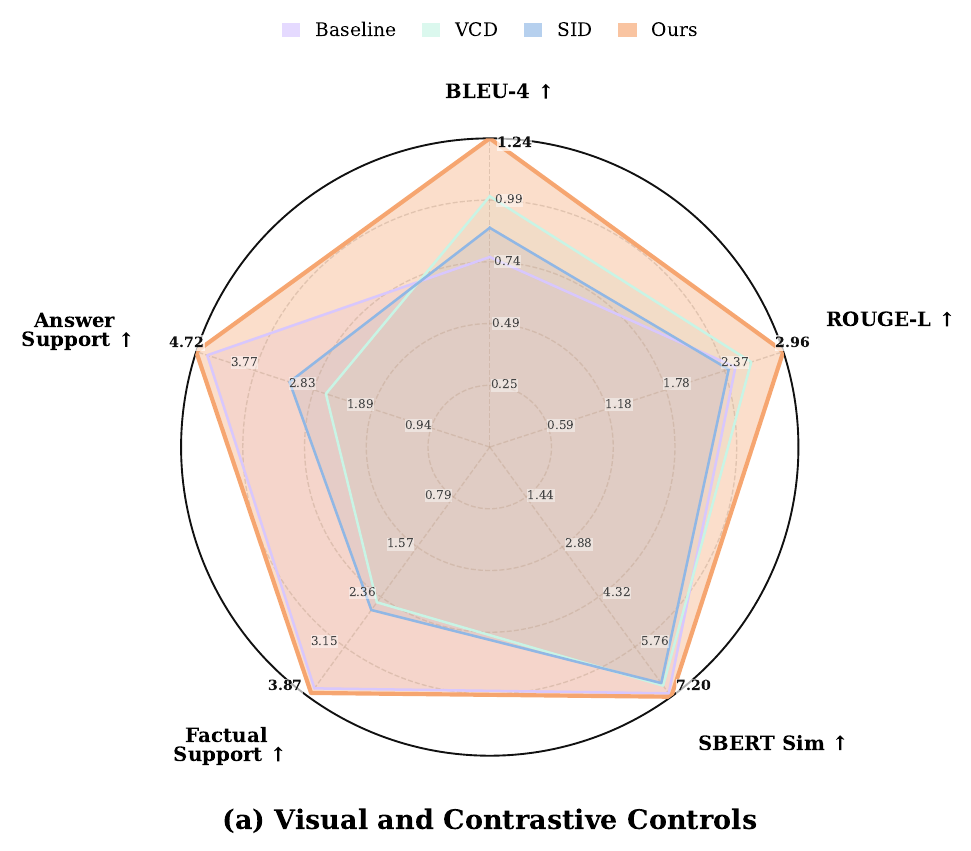}
\hfill
\includegraphics[width=0.49\columnwidth]{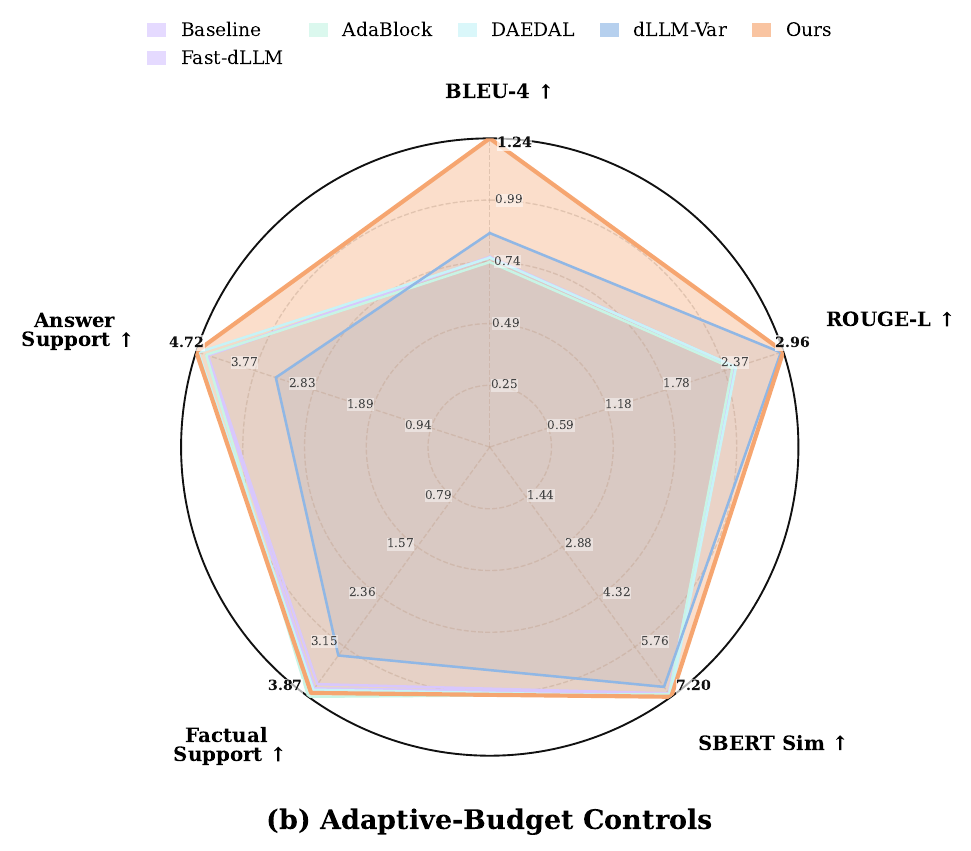}
\caption{Auxiliary CoT rationale-quality diagnostics on the ScienceQA-IMG active-control subset. 
The axes include BLEU-4, ROUGE-L, SBERT similarity, factual support, and answer support. 
The two judge dimensions correspond to Reference-Grounded Factuality and Rationale-Answer Alignment. 
The left panel compares visual and contrastive controls, while the right panel compares adaptive-budget controls. 
Both panels use the same zero-based per-axis normalization for visualization. 
These diagnostics are auxiliary and should not be interpreted as evidence of causal rationale faithfulness.}
\label{fig:scienceqa-cot-evaluation}
\end{figure}
\subsection{Length-Policy Stress Test}

A central question is whether the improvement comes only from making outputs shorter. 
The results do not support this explanation. 
Table~\ref{tab:length-policy-stress} compares representative settings where short, long, and routed decoding behave differently. 
On MME, Len2 is already strong, suggesting that many compact visual questions reach answer closure early. 
Routed control still slightly improves over Len2 while avoiding a global short policy. 
On ScienceQA-IMG, routed control improves over both Len2 and Len128, indicating that final-answer gains are not explained by uniformly shortening the output. 
On MME-CoT, Len2 is worse than Len128, while routed control improves over both. 
This shows that early commitment is unsafe when reasoning support is required.

\begin{table}[t]
\centering
\small
\setlength{\tabcolsep}{4pt}
\renewcommand{\arraystretch}{1.08}
\begin{tabular}{lcccc}
\toprule
Setting & Len2 & Len128 & Ours & Steps \\
\midrule
MME & 79.56 & 73.41 & \textbf{79.96} & 128.00 \\
SQA-IMG & 84.28 & 75.31 & \textbf{88.60} & 118.09 \\
MME-CoT & 48.72 & 49.57 & \textbf{52.42} & 102.09 \\
\bottomrule
\end{tabular}
\caption{Length-policy stress test. Scores are accuracies, and Steps reports the average decoding steps of routed control. The pattern shows that routed control is not equivalent to uniformly shortening the output.}
\label{tab:length-policy-stress}
\end{table}

Table~\ref{tab:route-distribution-main} gives a complementary route-level diagnostic. 
If routed control simply behaved like short decoding, most examples would be routed to commitment across all settings. 
Instead, the allocation changes with the evaluation interface and trajectory state. 
MME is dominated by commitment, which matches its compact answer-focused interface. 
In ScienceQA-IMG and MME-CoT diagnostic CoT settings, premature commitment is suppressed, and the controller mainly preserves the baseline or applies reasoning-supportive decoding.

\begin{table}[t]
\centering
\small
\setlength{\tabcolsep}{5pt}
\renewcommand{\arraystretch}{1.08}
\begin{tabular}{lccc}
\toprule
Setting & Commit & Preserve & Reason \\
\midrule
MME & 75.99 & 24.01 & 0.00 \\
SQA-IMG$^\dagger$ & 0.00 & 88.75 & 11.25 \\
MME-CoT & 0.00 & 71.79 & 28.21 \\
\bottomrule
\end{tabular}
\caption{Diagnostic route distribution across evaluation settings. Values are percentages. $^\dagger$ScienceQA-IMG uses the CoT diagnostic route interface.}
\label{tab:route-distribution-main}
\end{table}

Together, the two diagnostics clarify the role of routing. 
Short decoding helps when the answer is already closed, but it can remove useful intermediate structure in CoT-sensitive settings. 
Long decoding preserves reasoning capacity, but it can continue refining samples whose answer has already stabilized. 
Routed control improves robustness by selecting the safer action under the observed trajectory, rather than by committing to one global length policy.

\subsection{Auxiliary CoT Diagnostics on ScienceQA-IMG}

Final-answer accuracy does not show whether a generated reasoning trace resembles the reference explanation or supports the predicted answer. 
We therefore add an auxiliary CoT diagnostic on the ScienceQA-IMG active-control subset. 
For automatic evaluation, we compare each generated CoT trace with the ScienceQA reference explanation and report BLEU-4, ROUGE-L, and SBERT similarity. 
BLEU-1 is reported in Appendix as a supplementary overlap metric. 
For judge-based diagnostics, GPT-5.5 scores two text-only support dimensions on a 0--10 scale, factual support and answer support. 
The judge sees the question, reference explanation, predicted answer, and generated reasoning trace, but these scores are used only as auxiliary diagnostics.

Figure~\ref{fig:scienceqa-cot-evaluation} and Table~\ref{tab:scienceqa-cot-automatic} show that routed control improves reference-based explanation metrics over the fixed long-budget baseline. 
BLEU-4 increases from 0.105 to 0.124, ROUGE-L from 0.268 to 0.296, and SBERT similarity from 0.697 to 0.720. 
The paired tests are significant for all three metrics, with $p=1.03{\times}10^{-5}$ for BLEU-4, $p=8.53{\times}10^{-6}$ for ROUGE-L, and $p=3.41{\times}10^{-3}$ for SBERT similarity. 
The judge scores provide complementary evidence that routed control improves answer support while remaining competitive on factual support.

These results are auxiliary rather than definitive evidence of reasoning faithfulness. 
They show that routed control can improve actively controlled reasoning traces on a rationale-bearing benchmark, but they do not establish that the generated rationales faithfully explain the model's predictions.



\section{Conclusion}


Diffusion VLMs expose intermediate answer trajectories before the final output is fixed. 
This makes fixed-budget decoding a poor fit for questions with different closure times. 
Some examples already contain a stable, visually supported answer and can be harmed by continued refinement. 
Others require additional refinement, so early commitment can remove a useful intermediate structure.

We study this problem as reasoning-budget mismatch and address it with a parameter-frozen routed controller. 
The controller chooses among early commitment, fixed-budget preservation, and reasoning-supportive decoding from trajectory evidence rather than from a global length rule. 
Across answer-focused, mixed-reasoning, and CoT-sensitive settings, this routing view avoids the main failure modes of both pure short decoding and fixed long decoding. 
The broader implication is that diffusion VLM decoding should be treated as trajectory-aware control, not as universal generation-length selection.


\section*{Limitations}
This study has several limitations. First, the controller is evaluated on LLaDA-V, and additional diffusion VLMs are needed to establish model-level generality. Second, the routing signals are label-free proxies rather than correctness estimators. In particular, visual-attention evidence measures image-token engagement, not whether the answer is visually true. Third, some operating points use benchmark-specific guards or reasoning-supportive configurations; although these are fixed before evaluation, future work should study more unified threshold selection and calibration. Fourth, our CoT diagnostics combine reference-based automatic metrics with judge-based support scores, which should not be interpreted as causal evidence of rationale faithfulness. Finally, while routed control improves robustness, it is not primarily an acceleration method in the reported setting, since some routes retain long decoding budgets to preserve reasoning support.

\section*{Acknowledgments}
This study was supported by the National Natural Science Foundation of China (T2422012); the National Key Research and Development Program of China (2023YFC2415400); the Guangdong Basic and Applied Basic Research (2024B1515020088); the Shenzhen Science and Technology Program (ZDYJ20251211121037006); the High Level of Special Funds (G030230001, G03034K003); the Guangdong S\&T Program (2025B1111080001); the SUSTech Fang Keng Faculty Award.


\bibliography{custom}

\clearpage
\appendix

\section{Appendix}
\label{sec:appendix}

\subsection{Available Diagnostic Comparisons}

Table~\ref{tab:router-diagnostic-comparisons} reports action-level diagnostic comparisons for settings where corresponding artifacts are available. 
These rows are not intended as a full component ablation. 
They compare the full routed setting with available preserve-only, commit-only, separated-source, and guarded-routing variants.

The comparisons show two patterns. 
First, the full router improves over preserve-only baselines across all reported settings. 
Second, the full router is not equivalent to a commit-only policy. 
Commit proxy is competitive on answer-focused settings, but it underperforms the full router and drops on MME-CoT, where premature commitment can remove reasoning support. 
The ScienceQA-IMG CoT separated-source setting reaches the same answer accuracy as the full routed setting, suggesting that answer accuracy alone does not capture the effect of source selection in this diagnostic protocol. 
On MME-CoT, the guarded route improves over the unguarded route variant, indicating that the final guard helps reduce harmful route-on cases. 
We treat these rows as diagnostic action comparisons, while full component isolation is left to future controlled ablations.

\subsection{Routing Guards and Operating Points}
\label{app:routing-config}

$K(x)$ is an interface-level CoT-sensitive guard. 
It is enabled when the benchmark prompt requests a rationale, the evaluation interface scores reasoning support, or a disclosed route configuration marks the input as CoT-sensitive. 
$G_{\mathrm{on}}(x)$ is the fixed route-on guard for the reported operating point. 
$\chi_{\mathrm{commit}}(x)$ contains hard exclusions such as parser unavailability, operating-point constraints, or route-specific exclusion rules.

For MME-CoT, $\lambda_{\mathrm{on}}$ corresponds to the answer-first CoT prompt with \texttt{jsd\_adaptive\_throttle} and fixed route-on thresholds. 
When Route v3 is reported, it should be interpreted as a disclosed MME-CoT operating point rather than a benchmark-independent routing rule. 
Rows that use cached fixed-budget and route-on candidate outputs are marked as offline trace-composition analyses and are not used for online latency claims.

\subsection{Budget and Efficiency Analysis}

\EfficiencyBudgetTable

Table~\ref{tab:efficiency-budget} reports the budget controls. 
Fixed128 / Long128 decoding is computationally expensive across MME, ScienceQA-IMG, MME-CoT, and MMMU. 
Short2 is much cheaper and is often strong in answer-closed settings, reaching 79.56 on MME, 84.28 on ScienceQA-IMG, and 48.89 on MMMU with a much smaller decoding budget. 
This confirms that many answer-closed examples do not need the full fixed refinement path.

However, Short2 is not CoT-safe: on MME-CoT it reaches 48.72, below the Fixed128 CoT baseline at 49.57. Intermediate budgets also do not consistently solve the mismatch. Short32 and Fixed64 are below Short2 on MME and ScienceQA-IMG, and they do not improve over Fixed128 on MME-CoT. Route v3's MME-CoT gain is not explained by shorter outputs alone because its average steps and latency are comparable to the long CoT baseline. The method is therefore not simply reducing length; it selectively allocates or structures reasoning. These results motivate selective budget allocation rather than uniform budget reduction.

\begin{table}[t]
\centering
\small
\setlength{\tabcolsep}{3.6pt}
\renewcommand{\arraystretch}{1.08}
\begin{tabular}{lcccc}
\toprule
Variant & MME & MMMU & SQA-CoT & MME-CoT \\
\midrule
Full router 
& \textbf{79.96} 
& \textbf{49.44} 
& \textbf{88.60} 
& \textbf{52.42} \\
Preserve only 
& 73.41 
& 42.44 
& 84.18 
& 49.57 \\
Commit proxy 
& 79.56 
& 48.89 
& NA 
& 48.72 \\
Separated source 
& NA 
& NA 
& 88.60 
& NA \\
Unguarded route 
& NA 
& NA 
& NA 
& 51.57 \\
\bottomrule
\end{tabular}
\caption{Available diagnostic comparisons for routed decoding. Entries are accuracies. Preserve only denotes the fixed-budget or same-setting joint CoT baseline. Commit proxy uses the shortest benchmark-matched answer proxy. Separated source is reported only for ScienceQA-IMG CoT. Unguarded route is reported only for MME-CoT.}
\label{tab:router-diagnostic-comparisons}
\end{table}
\begin{table}[t]
\centering
\small
\setlength{\tabcolsep}{6.0pt}
\begin{tabular}{lcccc}
\toprule
Metric & Baseline & Method & Delta & $p$ \\
\midrule
BLEU-1    & 0.269 & 0.280 & +0.011 & $1.09{\times}10^{-4}$ \\
BLEU-4    & 0.105 & 0.124 & +0.018 & $1.03{\times}10^{-5}$ \\
ROUGE-L   & 0.268 & 0.296 & +0.028 & $8.53{\times}10^{-6}$ \\
SBERT sim & 0.697 & 0.720 & +0.023 & $3.41{\times}10^{-3}$ \\
\bottomrule
\end{tabular}
\caption{Automatic ScienceQA-IMG CoT explanation metrics on the active-control subset. BLEU-4, ROUGE-L, and SBERT similarity are used in Figure~\ref{fig:scienceqa-cot-evaluation}; BLEU-1 is reported as a supplementary overlap metric.}
\label{tab:scienceqa-cot-automatic}
\end{table}

\subsection{Shared Decoding Settings}

All reported rows use the frozen LLaDA-V model with the benchmark image input and task prompt assembled by the evaluation scripts.
We do not update model weights or train a verifier.
Unless a method-specific control requires a different field, decoding uses deterministic sampling with temperature 0, low-confidence remasking, and the benchmark parser used by the corresponding evaluation.

\begin{table}[t]
\centering
\small
\setlength{\tabcolsep}{4pt}
\renewcommand{\arraystretch}{1.08}
\begin{tabular}{ll}
\toprule
Field & Setting \\
\midrule
Model & Frozen LLaDA-V \\
Temperature & 0.0 \\
Remasking & Low confidence \\
Default budget & 128 steps, 128 generation length \\
Short budgets & 2, 32, and 64 steps \\
Parser & Original benchmark parser \\
Trace collection & Enabled for routed settings \\
\bottomrule
\end{tabular}
\caption{Shared decoding settings used in the reported experiments.}
\label{tab:shared-decoding-settings}
\end{table}

Fixed-budget rows keep the same model, prompt, and parser while changing the diffusion budget.
Len2, Len32, Len64, and Len128 use generation lengths matched to the corresponding number of denoising steps.
Adaptive-budget controls keep the same model and parser, while using their method-specific decoding rules.

\subsection{Routing Configuration Disclosure}

Table~\ref{tab:routing-configurations} summarizes the routing configuration used in each evaluation setting.
All routing rules are fixed before evaluation and do not use gold labels.
For answer-focused settings, the controller mainly decides whether the trajectory is safe to commit or should be preserved.
For CoT-sensitive settings, the reason route can be enabled when the interface requires support-bearing reasoning and the observed trajectory remains unresolved.

\begin{table}[t]
\centering
\small
\setlength{\tabcolsep}{3.5pt}
\renewcommand{\arraystretch}{1.08}
\begin{tabular}{lccc}
\toprule
Setting & Commit & Preserve & Reason \\
\midrule
ScienceQA-IMG & enabled & enabled & setting dependent \\
MME & enabled & enabled & disabled \\
MMMU & enabled & enabled & enabled \\
MME-CoT & disabled & enabled & enabled \\
\bottomrule
\end{tabular}
\caption{Routing actions enabled in each evaluation setting. ScienceQA-IMG uses a CoT diagnostic route interface for the rationale-quality analysis.}
\label{tab:routing-configurations}
\end{table}

Rows that use cached candidate outputs are used only for diagnostic composition and are not used for online latency claims.
This preserves the same benchmark prompt, parser, model weights, and branch-level decoding configuration, while separating diagnostic route analysis from online efficiency evaluation.

\subsection{LLM-Judge Prompt for ScienceQA-IMG CoT Diagnostics}
\label{app:llm-judge-prompt}

We use a text-only LLM judge as an auxiliary diagnostic for ScienceQA-IMG CoT outputs.
The judge does not inspect the image and does not see method names.
It receives the question, answer choices, reference explanation, predicted answer, and generated rationale.
The scores are used only to assess local support properties of the generated rationale and are not treated as evidence of causal rationale faithfulness.

\begin{promptbox}
You are evaluating a generated rationale for a ScienceQA-IMG multiple-choice example. This is a text-only evaluation. You cannot inspect the image. Use only the question, answer choices, reference explanation, predicted answer, and generated rationale. Do not infer unseen image contents. Do not use method names. Return JSON only.

Evaluate only the generated rationale. Metrics, each from 0 to 10:

1. factual_support: Does the rationale make claims that are supported by the question, answer choices, predicted answer, and reference explanation? A high score means the rationale is specific, plausible, and does not introduce unsupported facts.

2. answer_support: Does the rationale coherently support the predicted answer? A high score means the rationale explains why the predicted answer follows, rather than drifting toward another option or contradicting the predicted answer.

Scoring guide: 10 = accurate, specific, coherent, and well supported. 8 = mostly supported with minor omissions or harmless imprecision. 6 = partially supported but generic or incomplete. 4 = weak support, notable unsupported claims, or loose connection. 2 = mostly unsupported, contradictory, repetitive, or off-topic. 0 = empty, invalid, nonsensical, or supports a different answer.

Question: {question}
Answer choices: {choices}
Reference explanation: {reference_explanation}
Predicted answer: {predicted_answer}
Generated rationale: {generated_rationale}

Return exactly:
{
  "factual_support": integer from 0 to 10,
  "answer_support": integer from 0 to 10,
  "short_reason": "one short sentence"
}
\end{promptbox}

Judge calls use deterministic decoding with temperature 0.
The judge outputs are parsed as JSON.
We cache judge results by example and method to avoid repeated calls.
We do not request or store hidden chain-of-thought from the judge.
\subsection{Illustrative ScienceQA-IMG non-preserve CoT examples}
Illustrative ScienceQA-IMG non-preserve CoT examples (Case1--Case3). Each panel compares Len128 with Ours on the same example id and reports automatic explanation metrics and text-only judge scores. These diagnostics support rationale-quality analysis but do not prove faithfulness.
\begin{figure*}[t]
\centering
\includegraphics[width=0.98\textwidth]{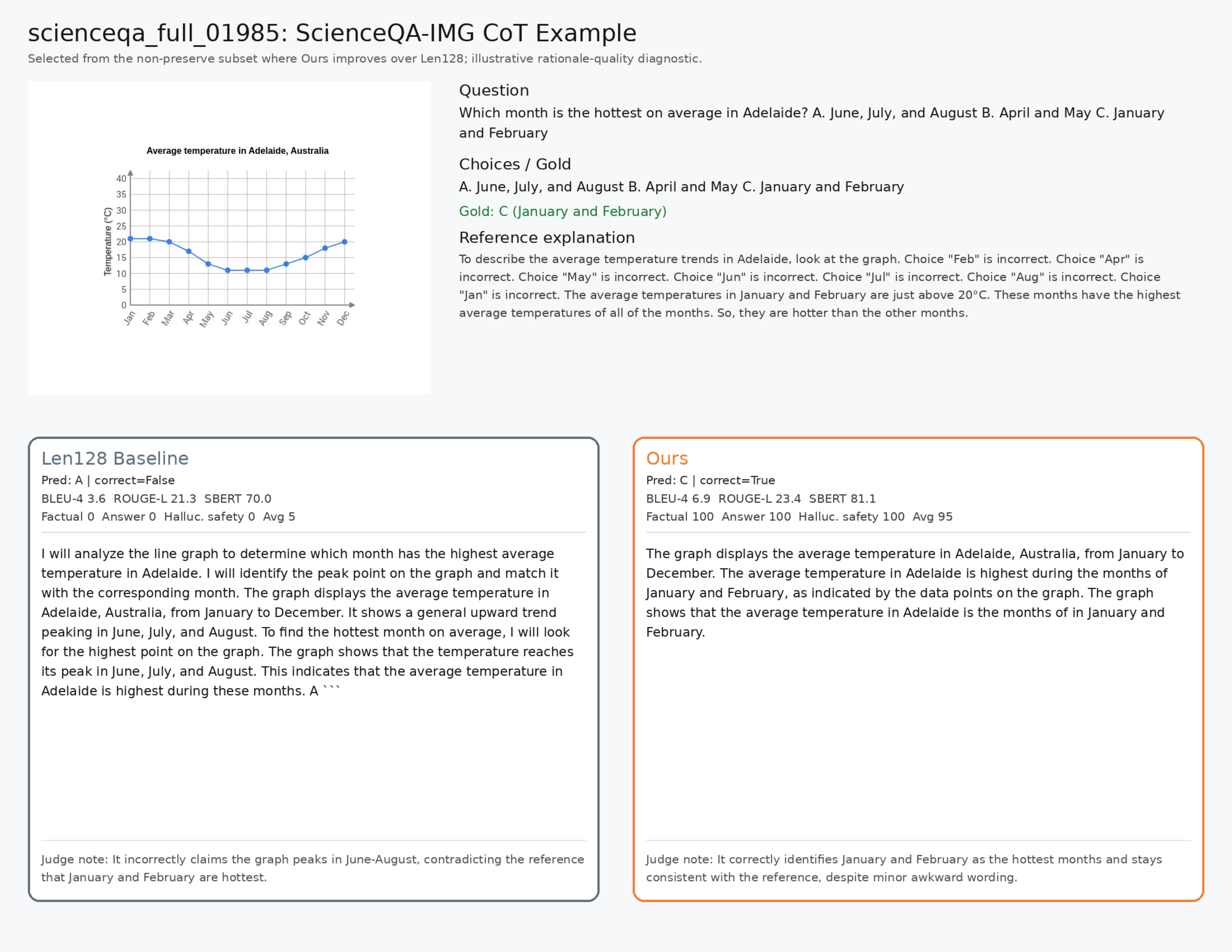}
\caption{Case1}
\end{figure*}

\begin{figure*}[t]
\centering
\includegraphics[width=0.98\textwidth]{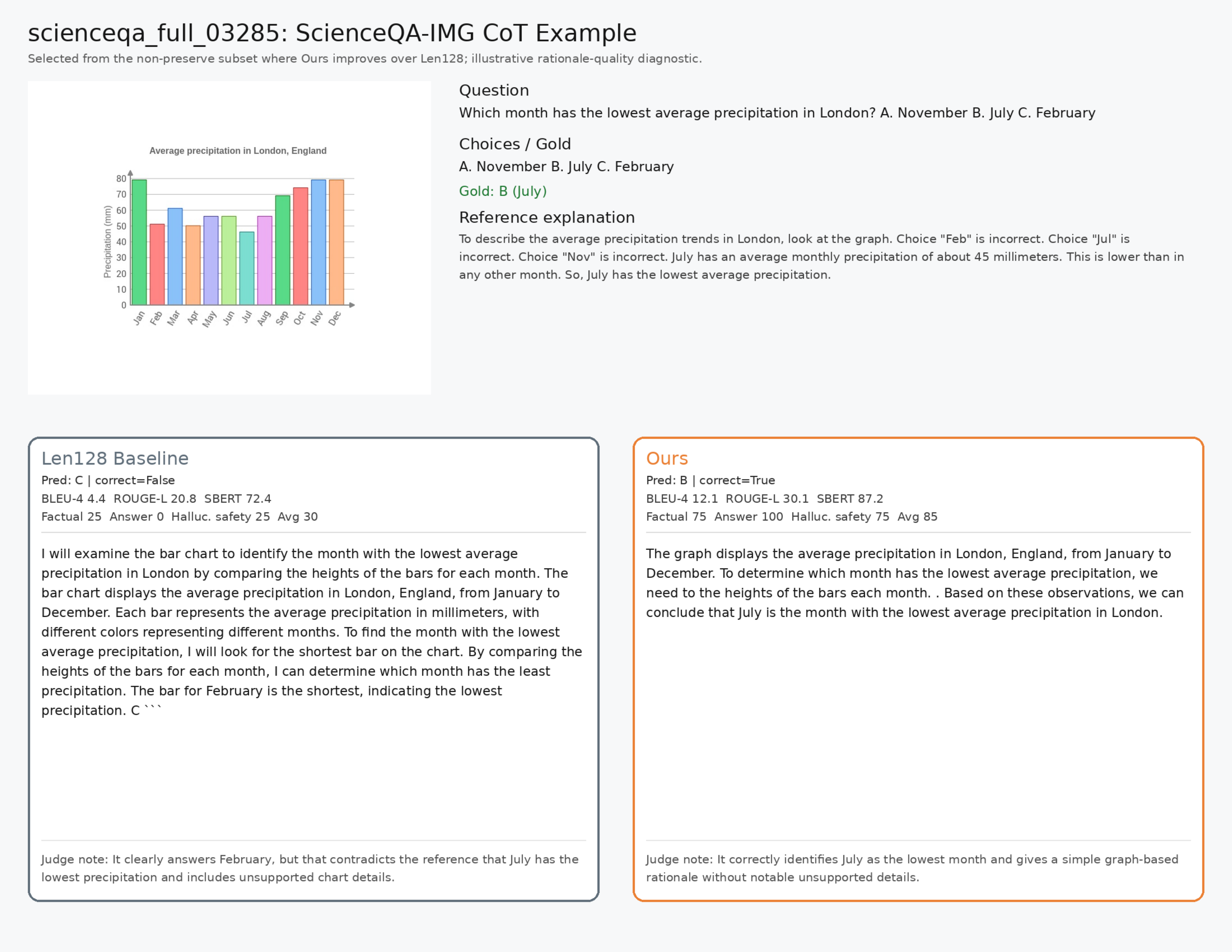}
\caption{Case2}
\end{figure*}

\begin{figure*}[t]
\centering
\includegraphics[width=0.98\textwidth]{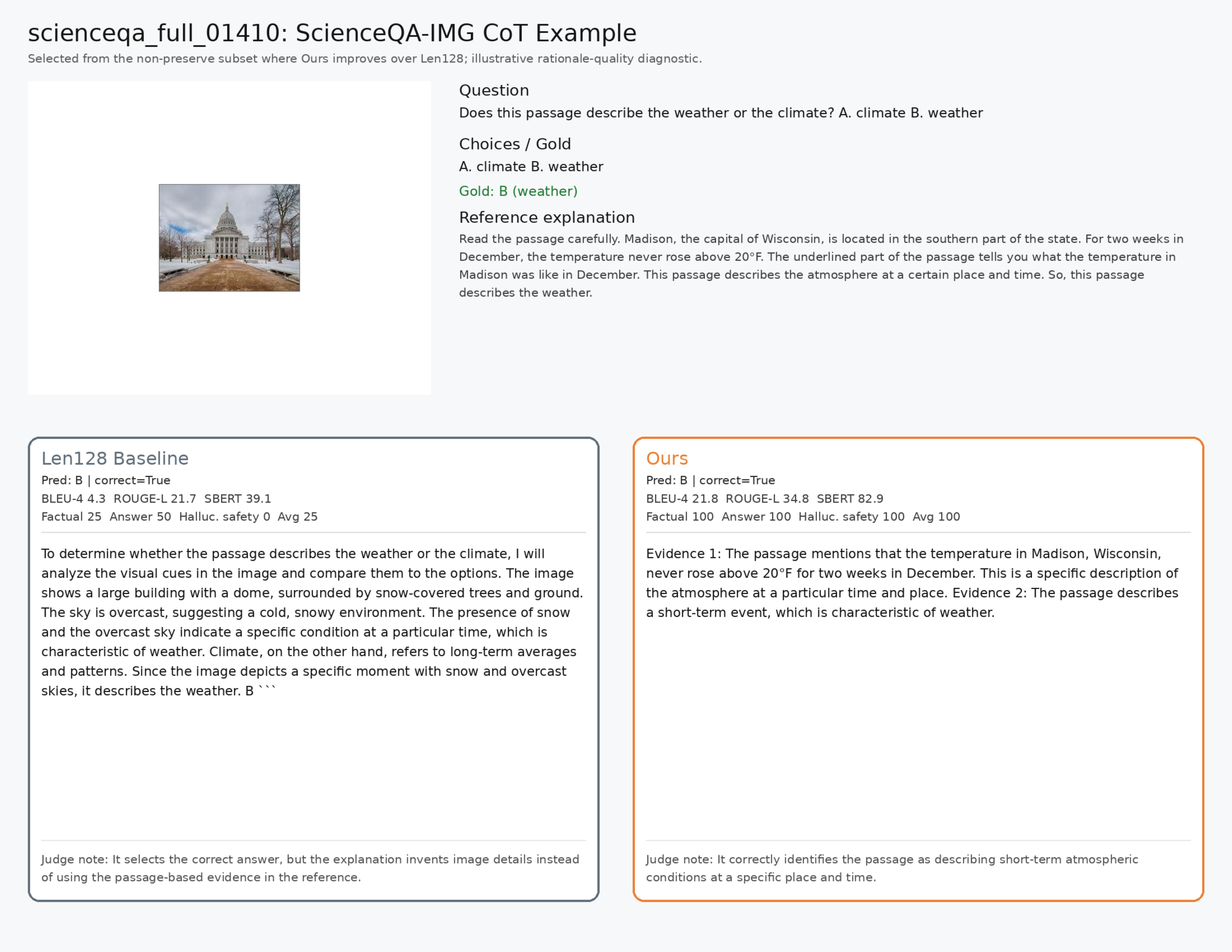}
\caption{Case3}
\end{figure*}

\subsection{Illustrative MME answer-token drift cases under fixed 128-step decoding.}
\begin{figure*}[t]
\centering
\includegraphics[width=0.98\textwidth]{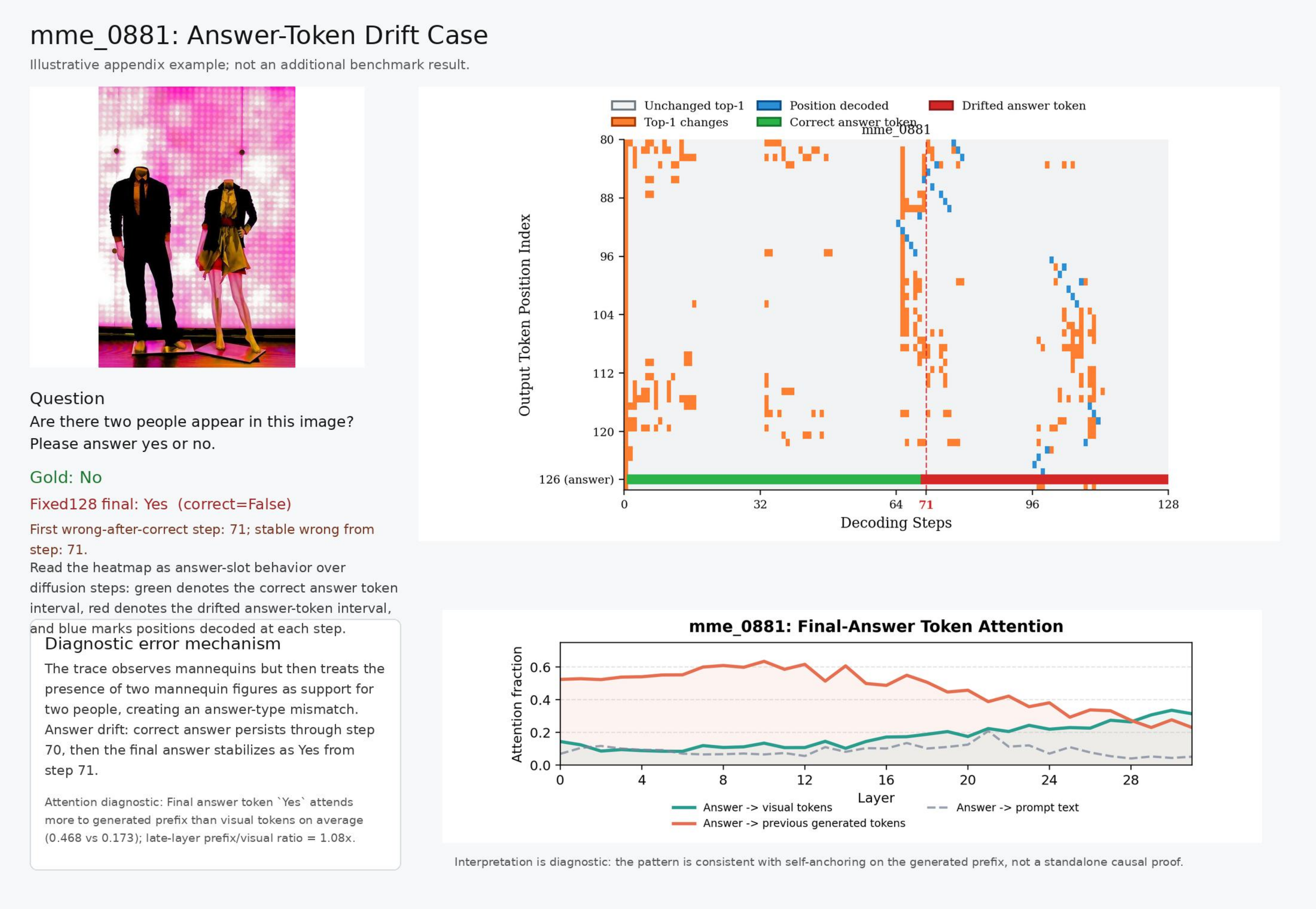}
\caption{Case4}
\end{figure*}

Illustrative MME answer-token drift cases (Case4 and Case 5) under fixed 128-step decoding. Green marks the correct answer-token interval, red marks the drifted answer-token interval, blue marks decoded positions, and orange marks top-1 changes. These examples diagnose answer-slot drift and do not constitute additional benchmark results.
\begin{figure*}[t]
\centering
\includegraphics[width=0.98\textwidth]{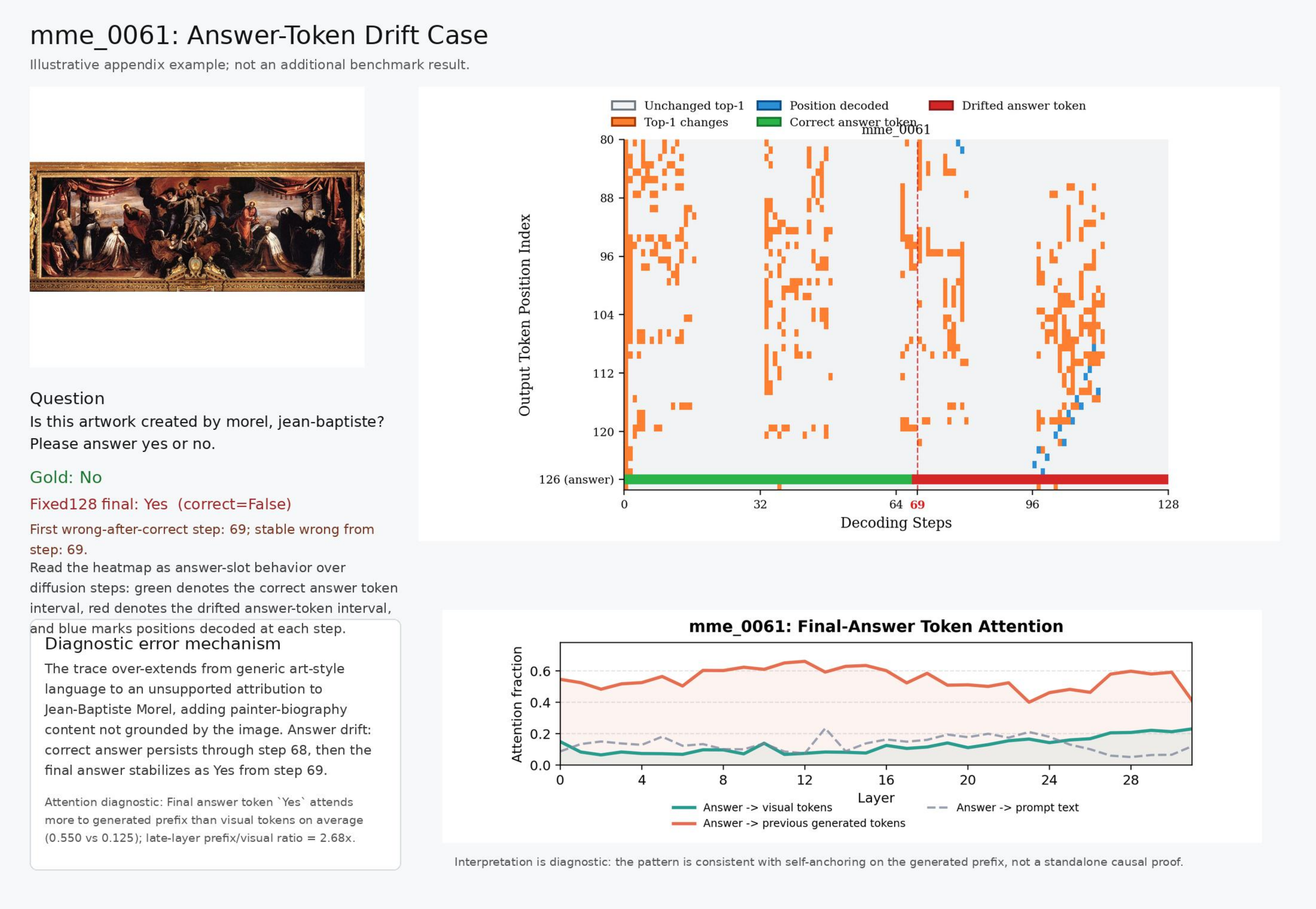}
\caption{Case5}
\end{figure*}
\end{document}